\documentclass[sigconf]{acmart}

\AtBeginDocument{%
  \providecommand\BibTeX{{%
    \normalfont B\kern-0.5em{\scshape i\kern-0.25em b}\kern-0.8em\TeX}}}

\copyrightyear{2019}
\acmYear{2019}
\setcopyright{acmlicensed}
\acmConference[Knoxville '19]{Knoxville '19: International Conference on Neuromorphic Systems}{July 23--25, 2019}{Knoxville, TN}
\acmBooktitle{2019 International Conference on Neuromorphic Systems, July 23--25, 2019, Knoxville, TN}

\usepackage{comment}
\begin{document}

%
\title[Neuromorphic Acceleration for Permanent Dropout]{Neuromorphic Acceleration for Approximate Bayesian Inference on Neural Networks via Permanent Dropout}

%

\author{Nathan Wycoff}
\authornote{To whom correspondence should be addressed.}
\authornote{Some of this work was conducted during an internship at the Mathematics and Computer Science Division of Argonne National Laboratory.}
\email{nathw95@vt.edu}
\affiliation{%
  \institution{Virginia Tech}
  \city{Blacksburg}
  \state{Virginia}
  \postcode{24060}
}
\author{Prasanna Balaprakash}
\email{pbalapra@anl.gov}
\author{Fangfang Xia}
\email{fangfang@anl.gov}
\affiliation{%
  \institution{Argonne National Laboratory}
  \streetaddress{9700 S. Cass Avenue}
  \city{Lemont}
  \state{Illinois}
  \postcode{60439}
}

%
\renewcommand{\shortauthors}{Wycoff, Balaprakash, and Xia}

%
\begin{abstract}
As neural networks have begun performing increasingly critical tasks for society, ranging from driving cars to identifying candidates for drug development, the value of their ability to perform uncertainty quantification (UQ) in their predictions has risen commensurately.
Permanent dropout, a popular method for neural network UQ, involves injecting stochasticity into the inference phase of the model and creating many predictions for each of the test data. This shifts the computational and energy burden of deep neural networks from the training phase to the inference phase. Recent work has demonstrated near-lossless conversion of classical deep neural networks to their spiking counterparts. We use these results to demonstrate the feasibility of conducting the inference phase with permanent dropout on spiking neural networks, mitigating the technique's computational and energy burden, which is essential for its use at scale or on edge platforms. We demonstrate the proposed approach via the Nengo spiking neural simulator on a combination drug therapy dataset for cancer treatment, where UQ is  critical. Our results indicate that the spiking approximation gives a predictive distribution practically indistinguishable from that given by the classical network. 
\end{abstract}

%

\keywords{Neuromorphic computing, Bayesian inference, Uncertainty quantification}

%
            \maketitle
\section{Introduction}

Deep neural networks (DNNs) are the arguable flagship of the machine learning (ML) revolution, having captured the imagination of the academic research community, industry, and to some extent the public at large because of their widespread empirical successes and captivating connection to human information processing. Historically sporting a black-box, predictive-error-driven approach, ML culture is increasingly interested in quantifying the uncertainty of its predictions. Standard, off-the-shelf tools from classical and Bayesian statistics to this end are often too computationally expensive to be of use in problems of even modest scale, a challenge the ML community has risen to meet. 

DNNs are increasingly being used for tasks that require quantification of prediction uncertainty. For instance, many autonomous vehicle frameworks are built on convolutional networks \cite{Janai2017}. Also, in the context of reinforcement learning with a DNN value function approximator,  understanding model uncertainty is important in order to determine where the agent should next explore \cite{Osband2016}. For camera relocalization, Kendall and Cipolla~\cite{Kendall2016} avail themselves of the uncertainty obtained from permadrop to obtain improvements in challenging indoor and outdoor problems. Recently, Thulasidasan et al.~\cite{Thulasidasan2019} developed a neural net with abstention, where the DNN may decide not to classify an instance if sufficient uncertainty exists. Furthermore, uncertainty quantification (UQ) is critical to many scientific ML applications as well \cite{Baker2018}. 

Dropout \cite{Srivastava2014}, an approach wherein individual neurons are randomly turned off (or otherwise perturbed), has been shown to be an effective approach for regularizing DNNs. 
The same approach applied during inference can approximate a Bayesian treatment of model uncertainty \cite{Gal2016}. In particular, it was shown that permanent dropout (called Monte Carlo dropout in the initial article and referred to as permadrop here) networks approximate a form of deep Gaussian processes \cite{Rasmussen2005,Damianou2013}. Traditionally, the cost of training dominates that of inference \cite{Goodfellow2016}; however, the permadrop strategy reverses this paradigm, since the inference phase must be executed many times, with increasing iterations giving increasing Monte Carlo accuracy. With the utility of UQ in DNNs having been carried out via permadrop, the challenge of reducing the concomitant computational and energy costs has become critical and nontrival. 


Spiking neural networks (SNNs) that run on neuromorphic hardware are a promising approach to address the computational and energy concerns of DNNs running on CPUs and GPUs for a class of applications. 
A recent study \cite{Blouw2018} using Intel Loihi \cite{Davies2018} found that it used 23.1 times fewer joules than a CPU (Xeon E5-2630) and 109.1 times fewer joules than a GPU (Quadro K4000) on an audio-processing problem with a two-layer neural net during the inference phase.
In this paper, we explore the prospect of offsetting the energy expense of the permadrop procedure in DNNs by converting them to SNNs during the inference phase. To do so, we expand upon the \texttt{nengo} and \texttt{nengo\_extras} \cite{Bekolay2014} packages, which allow conversion of simple DNNs to SNNs, implementing permadrop layers in the Nengo framework and demonstrating the feasibility of the process using the simulator therein.

\begin{figure}
    \centering
    \includegraphics[scale=0.4]{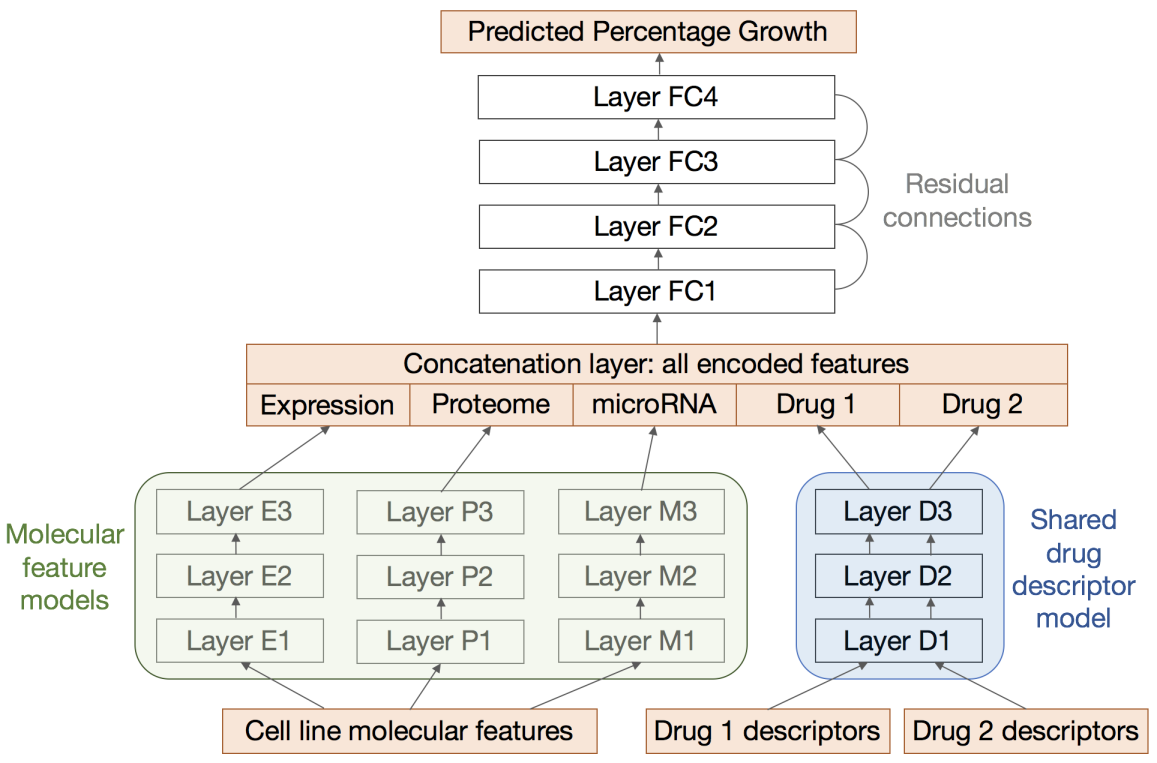}
    \caption{Architecture of the Combo neural network. Given two drugs, the aim is to predict the percent growth in human-derived cancer cell lines where these two drugs to be applied in a combination therapy.}
    \label{fig:combo}
\end{figure}

\section{Background and Related Work}
This article addresses a topic at the confluence of two threads of research: UQ on DNNs, and spiking conversion of DNNs.
\subsection{Permanent Dropout}

Dropout \cite{Srivastava2014} is a method for regularization in DNNs. In its simplest form, it involves randomly turning off neurons during each minibatch of training independently with some probability $p$. As originally proposed, the inference phase is unmodified aside from a scaling of the weights of each layer (as there are now more units present than during training). The intuition behind the method is that nodes cannot rely on a particular upstream or downstream neuron to modify their output and must instead pass on information that is more generally useful, as well as being forced to learn redundant representations. As outlined in \cite{Srivastava2014}, dropout may be viewed as approximate model averaging over all networks formed by subsets of the full network architecture. Gal and Ghahramani~\cite{Gal2016} showed that keeping dropout active during prediction (permadropout) is an approximation to a fully Bayesian treatment using a connection between neural networks and Gaussian processes. Each forward evaluation gives a random output; many forward evaluations build up an approximate predictive distribution.

\subsection{Spiking Conversion of Classical Neural Networks}

While SNNs are more powerful than DNNs in terms of theoretical computational ability \cite{Maass1997}, their often-discontinuous and computationally expensive nature means that training SNNs has been more challenging in practice than has been training DNNs, an already daunting task and the subject of major research. For this reason, the idea of conducting the training phase on a DNN and finding an SNN with similar behavior is an appealing one. Several approaches have been suggested for converting a DNN to an SNN while minimizing performance loss. Diehl et al.~\cite{Diehl2015} focused on converting DNNs with standard nonlinearities such as the softmax or ReLU functions, which was expanded upon by Rueckauer et al.~\cite{Rueckauer2017} to enable conversion of much more general neural architectures. 

Other work \cite{Cao2015} requires tailoring the DNN to optimize the SNN's performance, the approach we take in this paper. In particular, we follow the technique outlined in \cite{Hunsberger2015}, which simply requires using a specific activation function, termed the SoftLIF function.

Given a sufficiently large constant input to trigger an action potential, the firing rate of a linear Leaky Integrate and Fire (LIF) neuron with input current $\lambda$ is given by \cite{Gerstner2002}

\begin{equation}
    \frac{1}
    {\tau_{ref} + \tau_{RC} \log(1 + \frac{\nu}{\rho(\lambda - \nu)})},
\end{equation}
where $\rho(x) = \max[0,x]$. Unfortunately, this function is not continuously differentiable, complicating gradient-based optimization methods. To resolve this issue, Hunsberger and Eliasmith~\cite{Hunsberger2015} suggest replacing $\rho$ with a smooth approximation given by

\begin{equation}
    \sigma(x) = \gamma \log(1 + e^{\frac{x}{\gamma}}),
\end{equation}
which matches $\rho$ exactly as $\gamma\to 0$. Having trained a DNN with the SoftLIF activation, weights need simply to be transferred to an SNN of identical structure.

\begin{figure}
    \centering
    \includegraphics[scale=0.5]{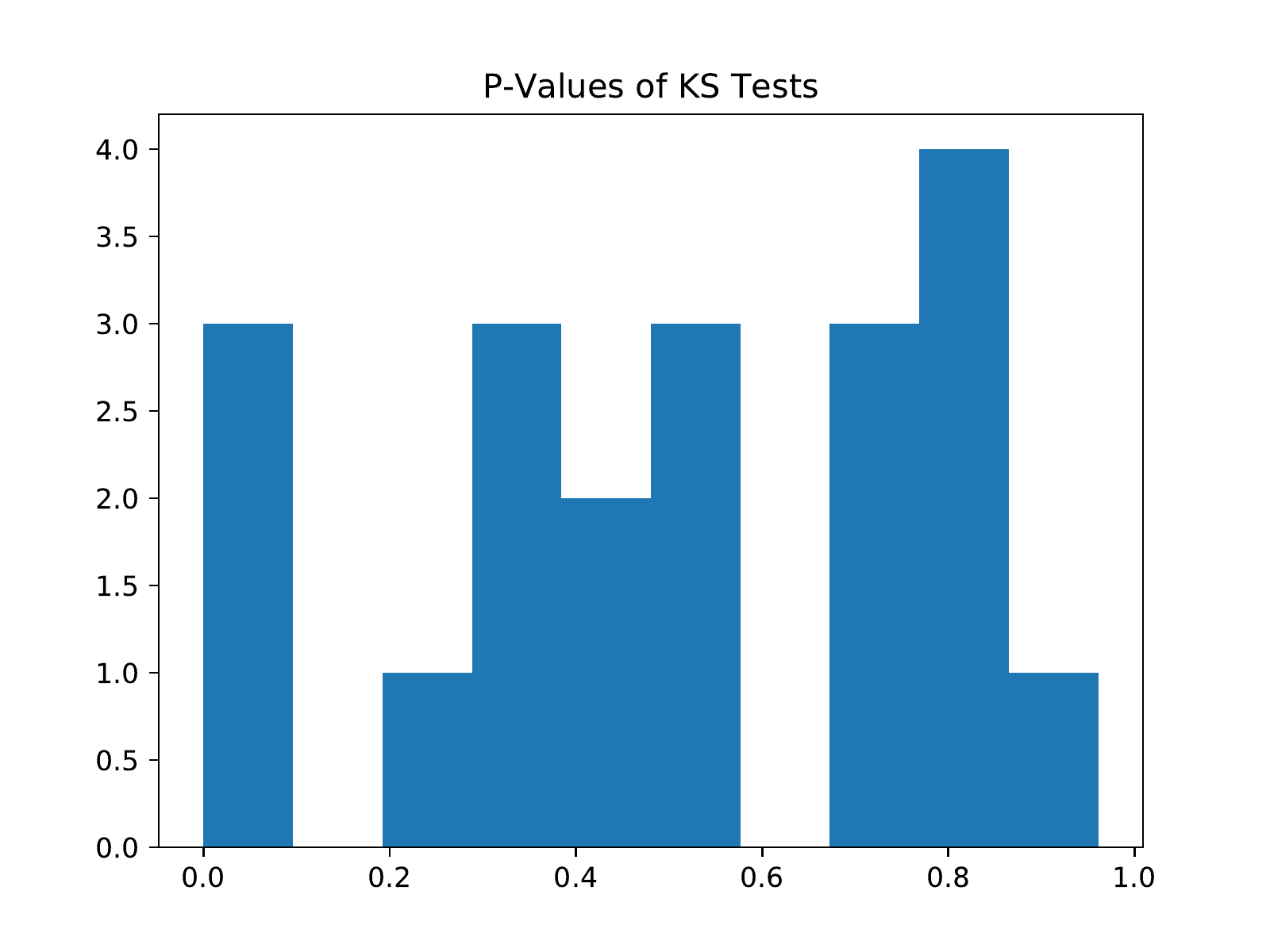}
    \caption{P-values for KS tests comparing samples of size 100 output by the DNN and the SNN for the first 20 observations. Were the distributions equal, we would expect the p-values to be approximately uniformly distributed.}
    \label{fig:kspvals}
\end{figure}

An SNN may be imbued with permadrop in a manner analogous to DNNs. We used the Nengo framework in simulations; since it  did not previously have support for permadrop, we modified the \texttt{nengo\_extras} package for our purposes. This task involved simply sampling a drop-mask for each layer during each simulation, that is, a binary vector of length equal to the number of neurons in a particular layer, in which 1's represent ``on" neurons, which will contribute to this simulation normally, and 0's represent ``off" neurons, which will not contribute at all. These vectors were sampled independently from a Bernoulli measure with some probability of success (i.e., neuron is active) $\rho$. A new drop-mask was sampled during each simulation, ultimately giving a distribution of outputs.

\section{Experimentation}

We executed our proposed method on the Combo benchmark of CANDLE, a U.S. Department of Energy Exascale Computing Project activity. The Combo deep neural network aims to predict the effectiveness of two drugs used in combination given tumor cell features (942 dimensions) as well as the description of each drug (3,820 dimensions), containing 248,656 observations. The data were obtained from the National Cancer Institute's ALMANAC resource \cite{Holbeck2017}. Network weights were shared for processing each drug of the pair; see Figure \ref{fig:combo} for details. In decision-making for cancer treatment, a complete accounting of uncertainty is critical, motivating the need for permadrop. On this benchmark, however, inference is expected to be 7 times more computationally expensive than training, because of UQ, underlining the potential gain from neuromorphic acceleration.

To implement our SNN, we used the Nengo framework \cite{Bekolay2014}, a Python-based spiking neuron simulator. Nengo allows conversion of feedforward neural networks implemented in, for instance, Keras \cite{Chollet2015}, into spiking Nengo objects, which may subsequently be simulated on a standard computer or in specialized hardware, such as Loihi. In our experiments, we trained a permadrop DNN using Keras with TensorFlow \cite{Tensorflow2015} as a backend.

\begin{figure}
    \centering
    \includegraphics[scale=0.5]{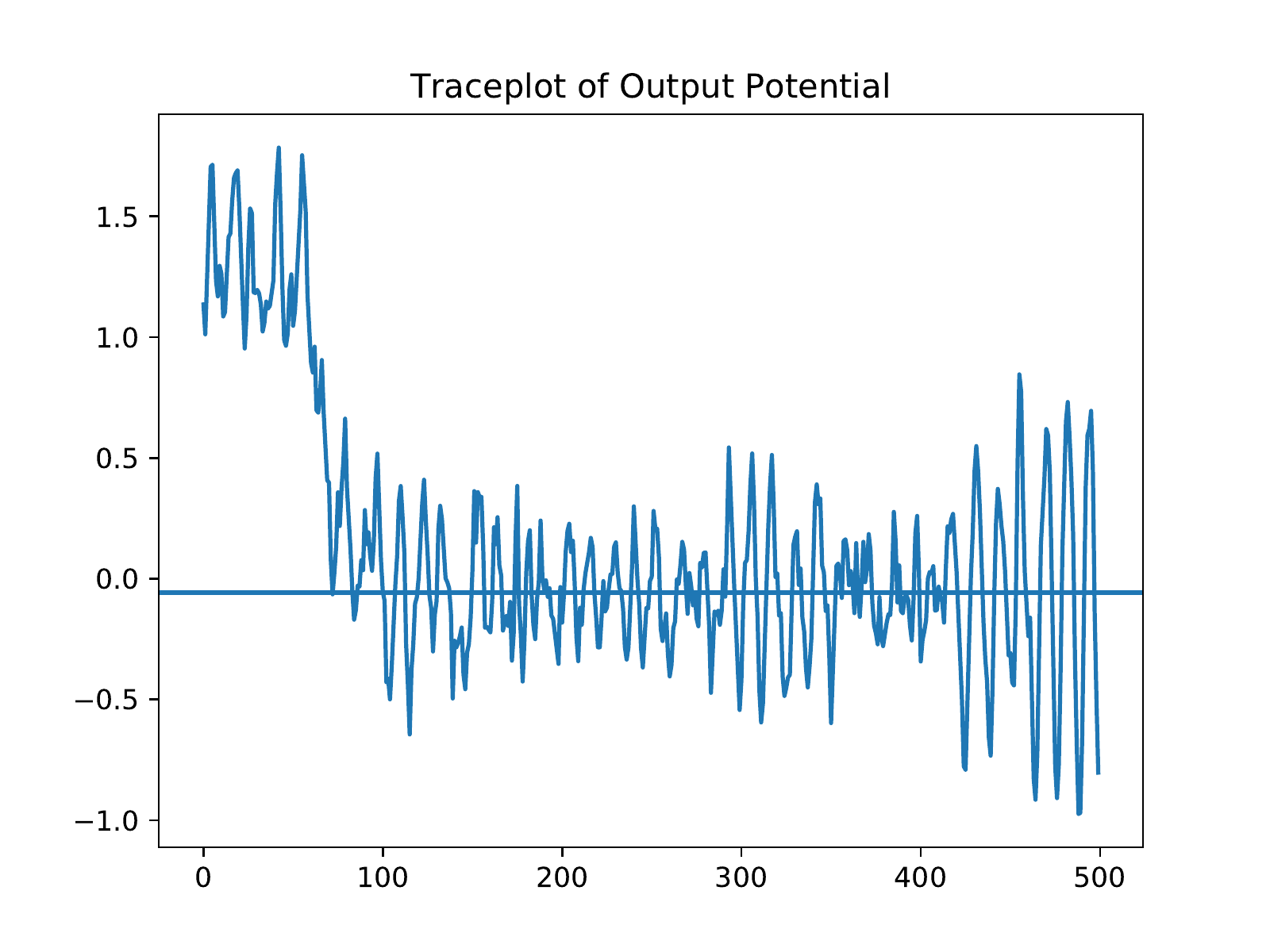}
    \caption{Output potential for the SNN on one observation. The transition time between states is removed using a 0.2 ms (or 200 tick) burn-in. Horizontal line gives DNN output.}
    \label{fig:trace}
\end{figure}

While the DNN's output is a scalar quantity giving predicted cell growth in percent, the output of its SNN analogue will be a time-valued quantity. We summarize the output potential over the time period by simply averaging the results, treating the first 0.2 ms as a "burn in" period and omitting the potential during this time from the average. Figure \ref{fig:trace} illustrates that the output potential of the SNN hovers around the output value of the DNN for most of the period on the first record of the Combo dataset. This same behavior is exhibited for all other observations. 

\begin{figure}
    \centering
    \includegraphics[scale=0.5]{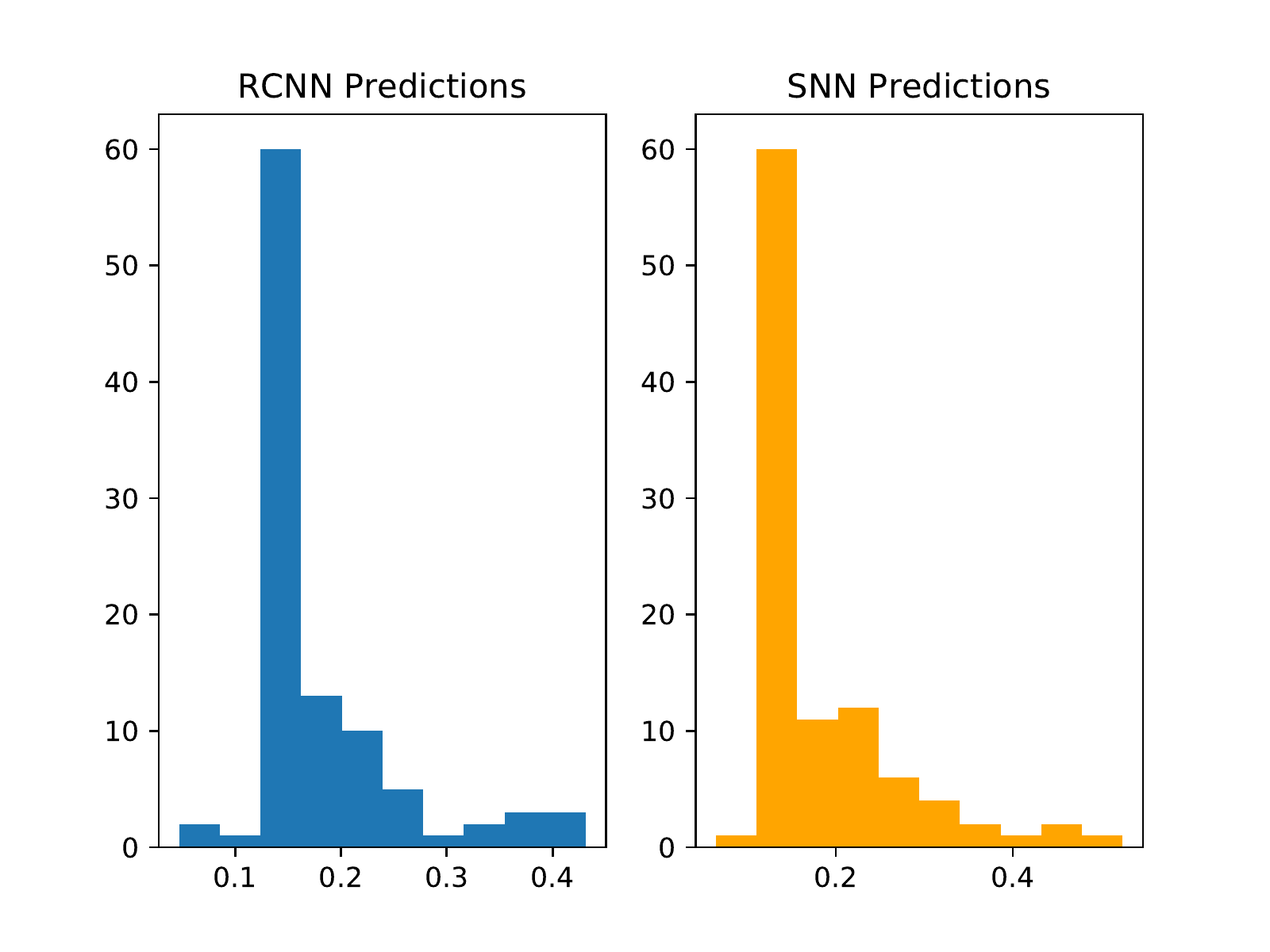}
    \caption{Histograms representing 100 draws from the predictive distributions for each neural network for the training example with the largest KS statistic (i.e., that with the most different distribution). Since they are similar visually, we conclude that the SNN is a good approximator.}
    \label{fig:ex_hist}
\end{figure}

We demonstrate that the distributions of outputs from the permadrop DNN and SNN are indistinguishable after averaging SNN output as described above after each dropout sampling. To quantitatively verify this claim, we got distributions of predictions for 100 observations containing 20 model forward steps each and ran a statistical hypothesis test that the two samples come from the same distribution. We used the Kolmogorov-Smirnov (KS) test, which involves measuring the infinity-norm difference (that is, maximum absolute discrepancy) between the empirical cumulative distribution functions of each sample. Figure \ref{fig:kspvals} gives a histogram of p-values from each pairwise comparison, corresponding to the output distributions of each neural net for a particular observation. In general hypothesis testing, under the null distribution, the p-value is uniformly distributed on the unit interval \cite{Casella2002}; however,  since the KS test is asymptotic, we should  expect this to hold only approximately in this case. We are satisfied that the KS test p-values generally seem to follow a uniform distribution,\footnote{A common criticism of KS tests (and general point-null hypothesis testing) is that for large sample sizes, even the smallest discrepancy will cause the test to reject the null hypothesis \cite{Wasserstein2016}. It is likely that we could consistently get results indicating that the two predictive distributions are different if we were willing to use a much larger sample size, though this would not mean that the distributions are, practically speaking, significantly different.} indicating that we could not detect a statistical difference between the two samples, and implying functional equivalence of the SNN and DNN. The histogram of the two predictive distributions corresponding to a single observation is shown in Figure \ref{fig:ex_hist} for illustration purposes.

\section{Conclusions and Perspectives}

We showed that permanent dropout for the purpose of approximate Bayesian predictive distribution computation on classical neural networks can be carried out on an SNN without any noticeable loss in distribution quality, opening the door for low-energy UQ via permadrop. We used the open source Nengo framework for simulation, which allows easy transfer of these models to neuromorphic hardware.

In our experiments, we first sampled a dropout mask, then ran an SNN with that mask, repeating this process many times to achieve a distribution of outputs. However, each of these outputs represents an aggregation of SNN potentials over some period of time. It may be possible to conduct the dropout sampling \textit{during} SNN simulation, such that the network connections are constantly changing in the SNN, and only one forward evaluation is required, even further reducing the computational burden. It is not \textit{a priori} clear whether the naive approach of simply sampling a different dropout mask at each iteration would match permadrop exactly or what modifications may be necessary. We leave investigations of such an approach to future work. 


All  this work was conducted on a simulator. A complete proof of concept would involve actual neuromorphic hardware and energy comparisons with standard DNNs run on standard hardware such as CPUs, GPUs, or TPUs.

%
\begin{acks}
N. Wycoff acknowledges funding from DOE LAB 17-1697 via a subaward from Argonne National Laboratory for SciDAC/DOE Office of Science ASCR and High Energy Physics. This material is based upon work supported by the U.S.\ Department of Energy 
(DOE), Office of Science, Office of Advanced Scientific Computing Research, under
Contract DE-AC02-06CH11357. 
\end{acks}

%
%
\bibliographystyle{ACM-Reference-Format}
\bibliography{sample-base}


\begin{thebibliography}{24}


\ifx \showCODEN    \undefined \def \showCODEN     #1{\unskip}     \fi
\ifx \showDOI      \undefined \def \showDOI       #1{#1}\fi
\ifx \showISBNx    \undefined \def \showISBNx     #1{\unskip}     \fi
\ifx \showISBNxiii \undefined \def \showISBNxiii  #1{\unskip}     \fi
\ifx \showISSN     \undefined \def \showISSN      #1{\unskip}     \fi
\ifx \showLCCN     \undefined \def \showLCCN      #1{\unskip}     \fi
\ifx \shownote     \undefined \def \shownote      #1{#1}          \fi
\ifx \showarticletitle \undefined \def \showarticletitle #1{#1}   \fi
\ifx \showURL      \undefined \def \showURL       {\relax}        \fi
\providecommand\bibfield[2]{#2}
\providecommand\bibinfo[2]{#2}
\providecommand\natexlab[1]{#1}
\providecommand\showeprint[2][]{arXiv:#2}

\bibitem[\protect\citeauthoryear{Abadi, Agarwal, Barham, Brevdo, Chen, Citro,
  Corrado, Davis, Dean, Devin, Ghemawat, Goodfellow, Harp, Irving, Isard, Jia,
  Jozefowicz, Kaiser, Kudlur, Levenberg, Man\'{e}, Monga, Moore, Murray, Olah,
  Schuster, Shlens, Steiner, Sutskever, Talwar, Tucker, Vanhoucke, Vasudevan,
  Vi\'{e}gas, Vinyals, Warden, Wattenberg, Wicke, Yu, and Zheng}{Abadi
  et~al\mbox{.}}{2015}]%
        {Tensorflow2015}
\bibfield{author}{\bibinfo{person}{Mart\'{\i}n Abadi}, \bibinfo{person}{Ashish
  Agarwal}, \bibinfo{person}{Paul Barham}, \bibinfo{person}{Eugene Brevdo},
  \bibinfo{person}{Zhifeng Chen}, \bibinfo{person}{Craig Citro},
  \bibinfo{person}{Greg~S. Corrado}, \bibinfo{person}{Andy Davis},
  \bibinfo{person}{Jeffrey Dean}, \bibinfo{person}{Matthieu Devin},
  \bibinfo{person}{Sanjay Ghemawat}, \bibinfo{person}{Ian Goodfellow},
  \bibinfo{person}{Andrew Harp}, \bibinfo{person}{Geoffrey Irving},
  \bibinfo{person}{Michael Isard}, \bibinfo{person}{Yangqing Jia},
  \bibinfo{person}{Rafal Jozefowicz}, \bibinfo{person}{Lukasz Kaiser},
  \bibinfo{person}{Manjunath Kudlur}, \bibinfo{person}{Josh Levenberg},
  \bibinfo{person}{Dandelion Man\'{e}}, \bibinfo{person}{Rajat Monga},
  \bibinfo{person}{Sherry Moore}, \bibinfo{person}{Derek Murray},
  \bibinfo{person}{Chris Olah}, \bibinfo{person}{Mike Schuster},
  \bibinfo{person}{Jonathon Shlens}, \bibinfo{person}{Benoit Steiner},
  \bibinfo{person}{Ilya Sutskever}, \bibinfo{person}{Kunal Talwar},
  \bibinfo{person}{Paul Tucker}, \bibinfo{person}{Vincent Vanhoucke},
  \bibinfo{person}{Vijay Vasudevan}, \bibinfo{person}{Fernanda Vi\'{e}gas},
  \bibinfo{person}{Oriol Vinyals}, \bibinfo{person}{Pete Warden},
  \bibinfo{person}{Martin Wattenberg}, \bibinfo{person}{Martin Wicke},
  \bibinfo{person}{Yuan Yu}, {and} \bibinfo{person}{Xiaoqiang Zheng}.}
  \bibinfo{year}{2015}\natexlab{}.
\newblock \bibinfo{title}{{TensorFlow}: Large-Scale Machine Learning on
  Heterogeneous Systems}.
\newblock
\newblock
\urldef\tempurl%
\url{https://www.tensorflow.org/}
\showURL{%
\tempurl}
\newblock
\shownote{Software available from tensorflow.org.}


\bibitem[\protect\citeauthoryear{Baker, Alexander, Bremer, Hagberg, Kevrekidis,
  Najm, Parashar, Patra, Sethian, Wild, and Willcox}{Baker
  et~al\mbox{.}}{2018}]%
        {Baker2018}
\bibfield{author}{\bibinfo{person}{Nathan Baker}, \bibinfo{person}{Frank
  Alexander}, \bibinfo{person}{Timo Bremer}, \bibinfo{person}{Aric Hagberg},
  \bibinfo{person}{Yannis Kevrekidis}, \bibinfo{person}{Habib Najm},
  \bibinfo{person}{Manish Parashar}, \bibinfo{person}{Abani Patra},
  \bibinfo{person}{James Sethian}, \bibinfo{person}{Stefan Wild}, {and}
  \bibinfo{person}{Karen Willcox}.} \bibinfo{year}{2018}\natexlab{}.
\newblock \showarticletitle{Brochure on Basic Research Needs for Scientific
  Machine Learning: Core Technologies for Artificial Intelligence}.
\newblock  (\bibinfo{date}{12} \bibinfo{year}{2018}).
\newblock
\urldef\tempurl%
\url{https://doi.org/10.2172/1484362}
\showDOI{\tempurl}


\bibitem[\protect\citeauthoryear{Bekolay, Bergstra, Hunsberger, DeWolf,
  Stewart, Rasmussen, Choo, Voelker, and Eliasmith}{Bekolay
  et~al\mbox{.}}{2014}]%
        {Bekolay2014}
\bibfield{author}{\bibinfo{person}{Trevor Bekolay}, \bibinfo{person}{James
  Bergstra}, \bibinfo{person}{Eric Hunsberger}, \bibinfo{person}{Travis
  DeWolf}, \bibinfo{person}{Terrence Stewart}, \bibinfo{person}{Daniel
  Rasmussen}, \bibinfo{person}{Xuan Choo}, \bibinfo{person}{Aaron Voelker},
  {and} \bibinfo{person}{Chris Eliasmith}.} \bibinfo{year}{2014}\natexlab{}.
\newblock \showarticletitle{Nengo: a Python tool for building large-scale
  functional brain models}.
\newblock \bibinfo{journal}{\emph{Frontiers in Neuroinformatics}}
  \bibinfo{volume}{7} (\bibinfo{year}{2014}), \bibinfo{pages}{48}.
\newblock
\showISSN{1662-5196}
\urldef\tempurl%
\url{https://doi.org/10.3389/fninf.2013.00048}
\showDOI{\tempurl}


\bibitem[\protect\citeauthoryear{Blouw, Choo, Hunsberger, and Eliasmith}{Blouw
  et~al\mbox{.}}{2018}]%
        {Blouw2018}
\bibfield{author}{\bibinfo{person}{Peter Blouw}, \bibinfo{person}{Xuan Choo},
  \bibinfo{person}{Eric Hunsberger}, {and} \bibinfo{person}{Chris Eliasmith}.}
  \bibinfo{year}{2018}\natexlab{}.
\newblock \showarticletitle{Benchmarking Keyword Spotting Efficiency on
  Neuromorphic Hardware}.
\newblock \bibinfo{journal}{\emph{CoRR}}  \bibinfo{volume}{abs/1812.01739}
  (\bibinfo{year}{2018}).
\newblock
\showeprint[arxiv]{1812.01739}
\urldef\tempurl%
\url{http://arxiv.org/abs/1812.01739}
\showURL{%
\tempurl}


\bibitem[\protect\citeauthoryear{Cao, Chen, and Khosla}{Cao
  et~al\mbox{.}}{2015}]%
        {Cao2015}
\bibfield{author}{\bibinfo{person}{Yongqiang Cao}, \bibinfo{person}{Yang Chen},
  {and} \bibinfo{person}{Deepak Khosla}.} \bibinfo{year}{2015}\natexlab{}.
\newblock \showarticletitle{Spiking Deep Convolutional Neural Networks for
  Energy-Efficient Object Recognition}.
\newblock \bibinfo{journal}{\emph{International Journal of Computer Vision}}
  \bibinfo{volume}{113} (\bibinfo{date}{05} \bibinfo{year}{2015}),
  \bibinfo{pages}{54--66}.
\newblock
\urldef\tempurl%
\url{https://doi.org/10.1007/s11263-014-0788-3}
\showDOI{\tempurl}


\bibitem[\protect\citeauthoryear{Casella and Berger}{Casella and
  Berger}{2002}]%
        {Casella2002}
\bibfield{author}{\bibinfo{person}{G. Casella} {and} \bibinfo{person}{R.L.
  Berger}.} \bibinfo{year}{2002}\natexlab{}.
\newblock \bibinfo{booktitle}{\emph{Statistical Inference}}.
\newblock \bibinfo{publisher}{Thomson Learning}.
\newblock
\showISBNx{9780534243128}
\showLCCN{2001025794}
\urldef\tempurl%
\url{https://books.google.com/books?id=0x\_vAAAAMAAJ}
\showURL{%
\tempurl}


\bibitem[\protect\citeauthoryear{Chollet et~al\mbox{.}}{Chollet
  et~al\mbox{.}}{2015}]%
        {Chollet2015}
\bibfield{author}{\bibinfo{person}{Fran\c{c}ois Chollet} {et~al\mbox{.}}}
  \bibinfo{year}{2015}\natexlab{}.
\newblock \bibinfo{title}{Keras}.
\newblock \bibinfo{howpublished}{\url{https://keras.io}}.
\newblock


\bibitem[\protect\citeauthoryear{Damianou and Lawrence}{Damianou and
  Lawrence}{2013}]%
        {Damianou2013}
\bibfield{author}{\bibinfo{person}{Andreas Damianou} {and}
  \bibinfo{person}{Neil Lawrence}.} \bibinfo{year}{2013}\natexlab{}.
\newblock \showarticletitle{Deep Gaussian Processes}. In
  \bibinfo{booktitle}{\emph{Proceedings of the Sixteenth International
  Conference on Artificial Intelligence and Statistics}}
  \emph{(\bibinfo{series}{Proceedings of Machine Learning Research})},
  \bibfield{editor}{\bibinfo{person}{Carlos~M. Carvalho} {and}
  \bibinfo{person}{Pradeep Ravikumar}} (Eds.), Vol.~\bibinfo{volume}{31}.
  \bibinfo{publisher}{PMLR}, \bibinfo{address}{Scottsdale, Arizona, USA},
  \bibinfo{pages}{207--215}.
\newblock
\urldef\tempurl%
\url{http://proceedings.mlr.press/v31/damianou13a.html}
\showURL{%
\tempurl}


\bibitem[\protect\citeauthoryear{{Davies}, {Srinivasa}, {Lin}, {Chinya}, {Cao},
  {Choday}, {Dimou}, {Joshi}, {Imam}, {Jain}, {Liao}, {Lin}, {Lines}, {Liu},
  {Mathaikutty}, {McCoy}, {Paul}, {Tse}, {Venkataramanan}, {Weng}, {Wild},
  {Yang}, and {Wang}}{{Davies} et~al\mbox{.}}{2018}]%
        {Davies2018}
\bibfield{author}{\bibinfo{person}{M. {Davies}}, \bibinfo{person}{N.
  {Srinivasa}}, \bibinfo{person}{T. {Lin}}, \bibinfo{person}{G. {Chinya}},
  \bibinfo{person}{Y. {Cao}}, \bibinfo{person}{S.~H. {Choday}},
  \bibinfo{person}{G. {Dimou}}, \bibinfo{person}{P. {Joshi}},
  \bibinfo{person}{N. {Imam}}, \bibinfo{person}{S. {Jain}}, \bibinfo{person}{Y.
  {Liao}}, \bibinfo{person}{C. {Lin}}, \bibinfo{person}{A. {Lines}},
  \bibinfo{person}{R. {Liu}}, \bibinfo{person}{D. {Mathaikutty}},
  \bibinfo{person}{S. {McCoy}}, \bibinfo{person}{A. {Paul}},
  \bibinfo{person}{J. {Tse}}, \bibinfo{person}{G. {Venkataramanan}},
  \bibinfo{person}{Y. {Weng}}, \bibinfo{person}{A. {Wild}}, \bibinfo{person}{Y.
  {Yang}}, {and} \bibinfo{person}{H. {Wang}}.} \bibinfo{year}{2018}\natexlab{}.
\newblock \showarticletitle{Loihi: A Neuromorphic Manycore Processor with
  On-Chip Learning}.
\newblock \bibinfo{journal}{\emph{IEEE Micro}} \bibinfo{volume}{38},
  \bibinfo{number}{1} (\bibinfo{date}{January} \bibinfo{year}{2018}),
  \bibinfo{pages}{82--99}.
\newblock
\showISSN{0272-1732}
\urldef\tempurl%
\url{https://doi.org/10.1109/MM.2018.112130359}
\showDOI{\tempurl}


\bibitem[\protect\citeauthoryear{Diehl, Neil, Binas, Cook, Liu, and
  Pfeiffer}{Diehl et~al\mbox{.}}{2015}]%
        {Diehl2015}
\bibfield{author}{\bibinfo{person}{Peter~U. Diehl}, \bibinfo{person}{Daniel
  Neil}, \bibinfo{person}{Jonathan Binas}, \bibinfo{person}{Matthew Cook},
  \bibinfo{person}{Shih-Chii Liu}, {and} \bibinfo{person}{Michael Pfeiffer}.}
  \bibinfo{year}{2015}\natexlab{}.
\newblock \showarticletitle{Fast-classifying, high-accuracy spiking deep
  networks through weight and threshold balancing}.
\newblock \bibinfo{journal}{\emph{2015 International Joint Conference on Neural
  Networks (IJCNN)}} (\bibinfo{year}{2015}), \bibinfo{pages}{1--8}.
\newblock


\bibitem[\protect\citeauthoryear{Gal and Ghahramani}{Gal and
  Ghahramani}{2016}]%
        {Gal2016}
\bibfield{author}{\bibinfo{person}{Yarin Gal} {and} \bibinfo{person}{Zoubin
  Ghahramani}.} \bibinfo{year}{2016}\natexlab{}.
\newblock \showarticletitle{Dropout As a Bayesian Approximation: Representing
  Model Uncertainty in Deep Learning}. In \bibinfo{booktitle}{\emph{Proceedings
  of the 33rd International Conference on International Conference on Machine
  Learning - Volume 48}} \emph{(\bibinfo{series}{ICML'16})}.
  \bibinfo{publisher}{JMLR.org}, \bibinfo{pages}{1050--1059}.
\newblock
\urldef\tempurl%
\url{http://dl.acm.org/citation.cfm?id=3045390.3045502}
\showURL{%
\tempurl}


\bibitem[\protect\citeauthoryear{Gerstner and Kistler}{Gerstner and
  Kistler}{2002}]%
        {Gerstner2002}
\bibfield{author}{\bibinfo{person}{Wulfram Gerstner} {and}
  \bibinfo{person}{Werner~M. Kistler}.} \bibinfo{year}{2002}\natexlab{}.
\newblock \bibinfo{booktitle}{\emph{Spiking Neuron Models: Single Neurons,
  Populations, Plasticity}}.
\newblock \bibinfo{publisher}{Cambridge University Press}.
\newblock
\urldef\tempurl%
\url{https://doi.org/10.1017/CBO9780511815706}
\showDOI{\tempurl}


\bibitem[\protect\citeauthoryear{Goodfellow, Bengio, and Courville}{Goodfellow
  et~al\mbox{.}}{2016}]%
        {Goodfellow2016}
\bibfield{author}{\bibinfo{person}{Ian Goodfellow}, \bibinfo{person}{Yoshua
  Bengio}, {and} \bibinfo{person}{Aaron Courville}.}
  \bibinfo{year}{2016}\natexlab{}.
\newblock \bibinfo{booktitle}{\emph{Deep Learning}}.
\newblock \bibinfo{publisher}{The MIT Press}.
\newblock
\showISBNx{0262035618, 9780262035613}


\bibitem[\protect\citeauthoryear{Holbeck, Camalier, Crowell, Govindharajulu,
  Hollingshead, Anderson, Polley, Rubinstein, Srivastava, Wilsker, Collins, and
  Doroshow}{Holbeck et~al\mbox{.}}{2017}]%
        {Holbeck2017}
\bibfield{author}{\bibinfo{person}{Susan~L. Holbeck}, \bibinfo{person}{Richard
  Camalier}, \bibinfo{person}{James~A. Crowell},
  \bibinfo{person}{Jeevan~Prasaad Govindharajulu}, \bibinfo{person}{Melinda
  Hollingshead}, \bibinfo{person}{Lawrence~W. Anderson}, \bibinfo{person}{Eric
  Polley}, \bibinfo{person}{Larry Rubinstein}, \bibinfo{person}{Apurva
  Srivastava}, \bibinfo{person}{Deborah Wilsker}, \bibinfo{person}{Jerry~M.
  Collins}, {and} \bibinfo{person}{James~H. Doroshow}.}
  \bibinfo{year}{2017}\natexlab{}.
\newblock \showarticletitle{The National Cancer Institute ALMANAC: A
  Comprehensive Screening Resource for the Detection of Anticancer Drug Pairs
  with Enhanced Therapeutic Activity}.
\newblock \bibinfo{journal}{\emph{Cancer Research}} \bibinfo{volume}{77},
  \bibinfo{number}{13} (\bibinfo{year}{2017}), \bibinfo{pages}{3564--3576}.
\newblock
\showISSN{0008-5472}
\urldef\tempurl%
\url{https://doi.org/10.1158/0008-5472.CAN-17-0489}
\showDOI{\tempurl}
\showeprint{http://cancerres.aacrjournals.org/content/77/13/3564.full.pdf}


\bibitem[\protect\citeauthoryear{Hunsberger and Eliasmith}{Hunsberger and
  Eliasmith}{2015}]%
        {Hunsberger2015}
\bibfield{author}{\bibinfo{person}{Eric Hunsberger} {and}
  \bibinfo{person}{Chris Eliasmith}.} \bibinfo{year}{2015}\natexlab{}.
\newblock \showarticletitle{Spiking Deep Networks with LIF Neurons}.
\newblock \bibinfo{journal}{\emph{CoRR}}  \bibinfo{volume}{abs/1510.08829}
  (\bibinfo{year}{2015}).
\newblock


\bibitem[\protect\citeauthoryear{{Janai}, {G{\"u}ney}, {Behl}, and
  {Geiger}}{{Janai} et~al\mbox{.}}{2017}]%
        {Janai2017}
\bibfield{author}{\bibinfo{person}{J. {Janai}}, \bibinfo{person}{F.
  {G{\"u}ney}}, \bibinfo{person}{A. {Behl}}, {and} \bibinfo{person}{A.
  {Geiger}}.} \bibinfo{year}{2017}\natexlab{}.
\newblock \showarticletitle{{Computer Vision for Autonomous Vehicles: Problems,
  Datasets and State-of-the-Art}}.
\newblock \bibinfo{journal}{\emph{arXiv e-prints}} (\bibinfo{date}{April}
  \bibinfo{year}{2017}).
\newblock
\showeprint[arxiv]{cs.CV/1704.05519}


\bibitem[\protect\citeauthoryear{{Kendall} and {Cipolla}}{{Kendall} and
  {Cipolla}}{2016}]%
        {Kendall2016}
\bibfield{author}{\bibinfo{person}{A. {Kendall}} {and} \bibinfo{person}{R.
  {Cipolla}}.} \bibinfo{year}{2016}\natexlab{}.
\newblock \showarticletitle{Modelling uncertainty in deep learning for camera
  relocalization}. In \bibinfo{booktitle}{\emph{2016 IEEE International
  Conference on Robotics and Automation (ICRA)}}. \bibinfo{pages}{4762--4769}.
\newblock
\urldef\tempurl%
\url{https://doi.org/10.1109/ICRA.2016.7487679}
\showDOI{\tempurl}


\bibitem[\protect\citeauthoryear{Maass}{Maass}{1997}]%
        {Maass1997}
\bibfield{author}{\bibinfo{person}{Wolfgang Maass}.}
  \bibinfo{year}{1997}\natexlab{}.
\newblock \showarticletitle{Networks of spiking neurons: The third generation
  of neural network models}.
\newblock \bibinfo{journal}{\emph{Neural Networks}} \bibinfo{volume}{10},
  \bibinfo{number}{9} (\bibinfo{year}{1997}), \bibinfo{pages}{1659 -- 1671}.
\newblock
\showISSN{0893-6080}
\urldef\tempurl%
\url{https://doi.org/10.1016/S0893-6080(97)00011-7}
\showDOI{\tempurl}


\bibitem[\protect\citeauthoryear{Osband, Blundell, Pritzel, and Van~Roy}{Osband
  et~al\mbox{.}}{2016}]%
        {Osband2016}
\bibfield{author}{\bibinfo{person}{Ian Osband}, \bibinfo{person}{Charles
  Blundell}, \bibinfo{person}{Alexander Pritzel}, {and}
  \bibinfo{person}{Benjamin Van~Roy}.} \bibinfo{year}{2016}\natexlab{}.
\newblock \showarticletitle{Deep Exploration via Bootstrapped DQN}.
\newblock In \bibinfo{booktitle}{\emph{Advances in Neural Information
  Processing Systems 29}}, \bibfield{editor}{\bibinfo{person}{D.~D. Lee},
  \bibinfo{person}{M.~Sugiyama}, \bibinfo{person}{U.~V. Luxburg},
  \bibinfo{person}{I.~Guyon}, {and} \bibinfo{person}{R.~Garnett}} (Eds.).
  \bibinfo{publisher}{Curran Associates, Inc.}, \bibinfo{pages}{4026--4034}.
\newblock
\urldef\tempurl%
\url{http://papers.nips.cc/paper/6501-deep-exploration-via-bootstrapped-dqn.pdf}
\showURL{%
\tempurl}


\bibitem[\protect\citeauthoryear{Rasmussen and Williams}{Rasmussen and
  Williams}{2005}]%
        {Rasmussen2005}
\bibfield{author}{\bibinfo{person}{Carl~Edward Rasmussen} {and}
  \bibinfo{person}{Christopher K.~I. Williams}.}
  \bibinfo{year}{2005}\natexlab{}.
\newblock \bibinfo{booktitle}{\emph{Gaussian Processes for Machine Learning
  (Adaptive Computation and Machine Learning)}}.
\newblock \bibinfo{publisher}{The MIT Press}.
\newblock
\showISBNx{026218253X}


\bibitem[\protect\citeauthoryear{Rueckauer, Lungu, Hu, Pfeiffer, and
  Liu}{Rueckauer et~al\mbox{.}}{2017}]%
        {Rueckauer2017}
\bibfield{author}{\bibinfo{person}{Bodo Rueckauer},
  \bibinfo{person}{Iulia-Alexandra Lungu}, \bibinfo{person}{Yuhuang Hu},
  \bibinfo{person}{Michael Pfeiffer}, {and} \bibinfo{person}{Shih-Chii Liu}.}
  \bibinfo{year}{2017}\natexlab{}.
\newblock \showarticletitle{Conversion of Continuous-Valued Deep Networks to
  Efficient Event-Driven Networks for Image Classification}.
\newblock \bibinfo{journal}{\emph{Front Neurosci}}  \bibinfo{volume}{11}
  (\bibinfo{date}{07 Dec} \bibinfo{year}{2017}), \bibinfo{pages}{682--682}.
\newblock
\showISSN{1662-4548}
\urldef\tempurl%
\url{https://doi.org/10.3389/fnins.2017.00682}
\showDOI{\tempurl}
\newblock
\shownote{29375284[pmid].}


\bibitem[\protect\citeauthoryear{Srivastava, Hinton, Krizhevsky, Sutskever, and
  Salakhutdinov}{Srivastava et~al\mbox{.}}{2014}]%
        {Srivastava2014}
\bibfield{author}{\bibinfo{person}{Nitish Srivastava},
  \bibinfo{person}{Geoffrey Hinton}, \bibinfo{person}{Alex Krizhevsky},
  \bibinfo{person}{Ilya Sutskever}, {and} \bibinfo{person}{Ruslan
  Salakhutdinov}.} \bibinfo{year}{2014}\natexlab{}.
\newblock \showarticletitle{Dropout: A Simple Way to Prevent Neural Networks
  from Overfitting}.
\newblock \bibinfo{journal}{\emph{Journal of Machine Learning Research}}
  \bibinfo{volume}{15} (\bibinfo{year}{2014}), \bibinfo{pages}{1929--1958}.
\newblock
\urldef\tempurl%
\url{http://jmlr.org/papers/v15/srivastava14a.html}
\showURL{%
\tempurl}


\bibitem[\protect\citeauthoryear{Thulasidasan, Bhattacharya, Bilmes,
  Chennupati, and Mohd-Yusof}{Thulasidasan et~al\mbox{.}}{2019}]%
        {Thulasidasan2019}
\bibfield{author}{\bibinfo{person}{Sunil Thulasidasan}, \bibinfo{person}{Tanmoy
  Bhattacharya}, \bibinfo{person}{Jeffrey Bilmes}, \bibinfo{person}{Gopinath
  Chennupati}, {and} \bibinfo{person}{Jamal Mohd-Yusof}.}
  \bibinfo{year}{2019}\natexlab{}.
\newblock \bibinfo{title}{Knows When it Doesn’t Know: Deep Abstaining
  Classifiers}.
\newblock
\newblock
\urldef\tempurl%
\url{https://openreview.net/forum?id=rJxF73R9tX}
\showURL{%
\tempurl}


\bibitem[\protect\citeauthoryear{Wasserstein and Lazar}{Wasserstein and
  Lazar}{2016}]%
        {Wasserstein2016}
\bibfield{author}{\bibinfo{person}{Ronald~L. Wasserstein} {and}
  \bibinfo{person}{Nicole~A. Lazar}.} \bibinfo{year}{2016}\natexlab{}.
\newblock \showarticletitle{The ASA's Statement on p-Values: Context, Process,
  and Purpose}.
\newblock \bibinfo{journal}{\emph{The American Statistician}}
  \bibinfo{volume}{70}, \bibinfo{number}{2} (\bibinfo{year}{2016}),
  \bibinfo{pages}{129--133}.
\newblock
\urldef\tempurl%
\url{https://doi.org/10.1080/00031305.2016.1154108}
\showDOI{\tempurl}
\showeprint{https://doi.org/10.1080/00031305.2016.1154108}


\end{thebibliography}

\begin{center}
    \framebox{\parbox{3.2in}{
    The submitted manuscript has been created by UChicago Argonne, LLC, Operator of Argonne National Laboratory (``Argonne"). Argonne, a U.S. Department of Energy Office of Science laboratory, is operated under Contract No. DE-AC02-06CH11357. The U.S. Government retains for itself, and others acting on its behalf, a paid-up nonexclusive, irrevocable worldwide license in said article to reproduce, prepare derivative works, distribute copies to the public, and perform publicly and display publicly, by or on behalf of the Government. The Department of Energy will provide public access to these results of federally sponsored research in accordance with the DOE Public Access Plan. \url{http://energy.gov/downloads/doe-public-access-plan}}}
    \normalsize
\end{center}
\end{document}